\documentclass{ieeeaccess_arxiv}
\usepackage{cite}
\usepackage{amsmath,amssymb,amsfonts}
\usepackage{algorithmic}
\usepackage{graphicx}
\usepackage{textcomp}
\usepackage{comment}
\usepackage{amsmath}
\usepackage{amsfonts}
\usepackage{bm}
\usepackage{subfigure}
\usepackage{booktabs}
\usepackage{framed}

\def\BibTeX{{\rm B\kern-.05em{\sc i\kern-.025em b}\kern-.08em
    T\kern-.1667em\lower.7ex\hbox{E}\kern-.125emX}}
\begin{document}
\history{Date of publication xxxx 00, 0000, date of current version xxxx 00, 0000.}
\doi{xxxx/xxxx}

\title{What Shape Is Optimal for Masks in Text Removal?}
\author{\uppercase{Hyakka Nakada}\authorrefmark{1}
\uppercase{and Marika Kubota}.\authorrefmark{2}}
\address[1]{Recruit Co., Ltd., Tokyo 100-6640, Japan}
\address[2]{Beans Labo Co., Ltd., Naha-shi, Okinawa 900-0006, Japan}

\onecolumn
\begin{framed}
   \noindent
   This work has been submitted to the IEEE for possible publication. Copyright may be transferred without notice, after which this version may no longer be accessible.
\end{framed}
\clearpage
\twocolumn

\markboth
{Author \headeretal: Preparation of Papers for xxxx TRANSACTIONS and JOURNALS}
{Author \headeretal: Preparation of Papers for xxxx TRANSACTIONS and JOURNALS}

\corresp{Corresponding author: Hyakka Nakada (e-mail: hyakka\_nakada@r.recruit.co.jp).}

\begin{abstract}
The advent of generative models has dramatically improved the accuracy of image inpainting. In particular, by removing specific text from document images, reconstructing original images is extremely important for industrial applications. However, most existing methods of text removal focus on deleting simple scene text which appears in images captured by a camera in an outdoor environment. There is little research dedicated to complex and practical images with dense text. 
Therefore, we created benchmark data for text removal from images including a large amount of text. From the data, we found that text-removal performance becomes vulnerable against mask profile perturbation.
Thus, for practical text-removal tasks, precise tuning of the mask shape is essential. This study developed a method to model highly flexible mask profiles and learn their parameters using Bayesian optimization. 
The resulting profiles were found to be character-wise masks. It was also found that the minimum cover of a text region is not optimal. 
Our research is expected to pave the way for a user-friendly guideline for manual masking.
\end{abstract}


\titlepgskip=-15pt

\maketitle

\section{INTRODUCTION}
\label{sec:introduction}
\PARstart{I}{mage} inpainting is a fundamental problem in computer vision, which aims to synthesize plausible and realistic content in the missing regions of images~\cite{inpaint_elharrouss, inpaint_pushpalwar, inpaint_jam, inpaint_qin, inpaint_xu}. 
There is a wide variety of applications, such as the restoration of old photographs and removal of unwanted objects.
With the advance of deep learning, inpainting problems have been effectively solved~\cite{inpaint_qin, inpaint_xu}.
In the last few years, Generative AI such as diffusion models has achieved significant progress in the overall capability of image generation~\cite{inpaint_lugmayr, inpaint_ju}.

In the field of image inpainting, the process begins with masking.
After original images are given, users define the corresponding mask. 
Typically, the mask is a binary image where pixels to be removed are marked to $1$ bit, and valid pixels are unmarked ($0$ bit).
An input to an inpainting network is the corrupted image, which is obtained by performing an AND operation between the original images and the mask.
The inpainting network is trained to predict the missing pixel values.
This network usually uses convolutional layers to extract features from the unmasked regions, such as patterns and contexts.
Thus, new pixel values can be synthesized within the masked region.
Finally, post-processing is sometimes applied, such as refinement of the boundaries~\cite{inpaint_edge} or super resolution~\cite{inpaint_super}.
\Figure[t!](topskip=0pt, botskip=0pt, midskip=0pt)[width=13.5cm]{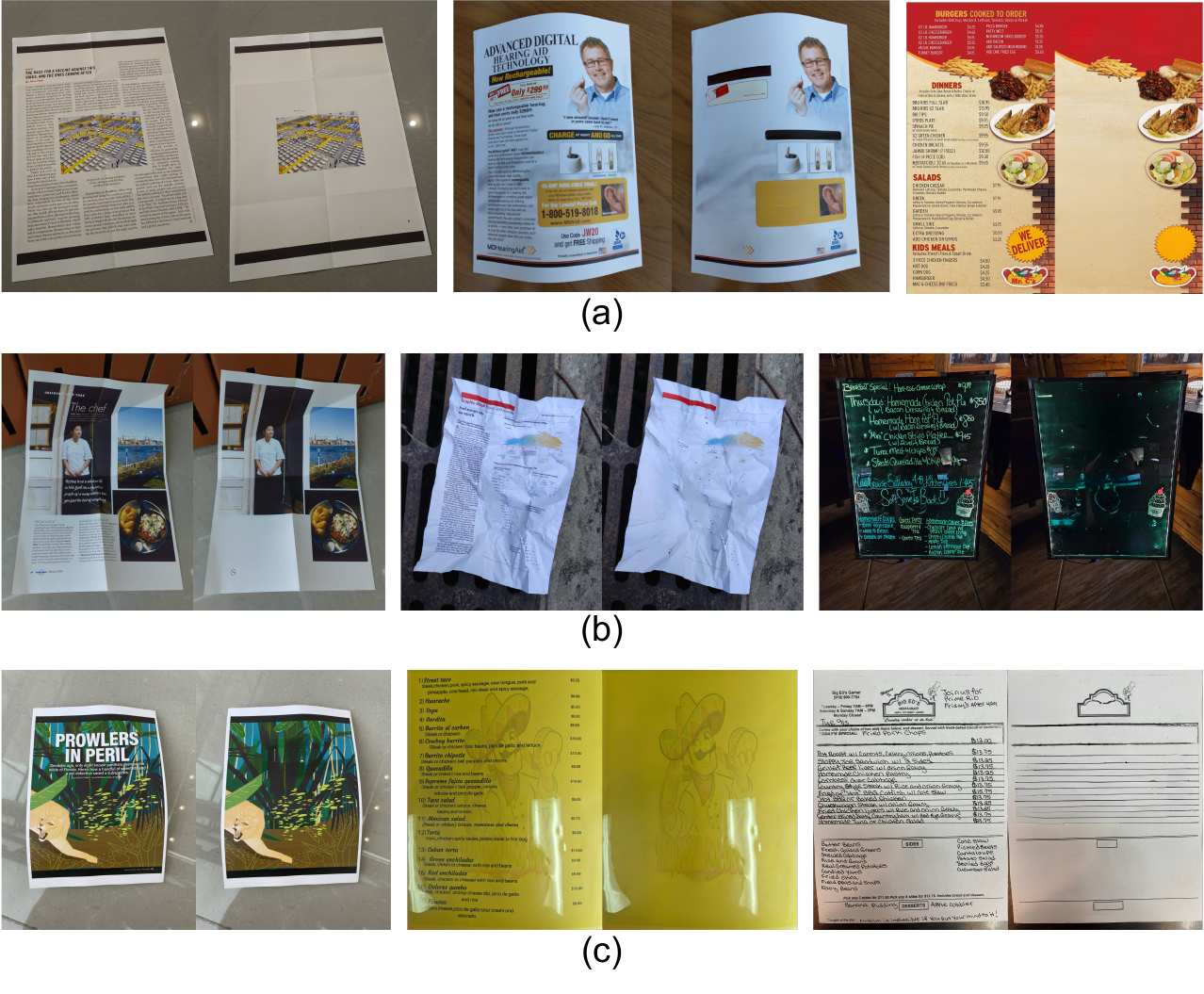}
{Example data of our benchmark. While inputs are original images, outputs are processed images, which are inpainted by text removal. As original images, we mainly collect document images with plentiful text (a). This data occasionally possesses high OCR difficulty due to perturbations such as folding, crumpling, and hand-writing (b). In addition, complex-background images are included (c).
\label{fig:benchmark_example}}

Masking process has the vast degrees of freedom.
Because masking is usually performed manually, the profile of masks may be prone to dependence on individuals.
Using a mask that is excessively broader than a target object to be removed will cause a significant deterioration in the accuracy of image inpainting. As an extreme example, masking the entire image makes restoration impossible.
Futhermore, it is not trivial that a mask with a minimum area that follows the object's contour is always preferable.
This is because when the contours are complex, the resulting mask may become irregular and different from existing training data.
Therefore, several efforts have reported to train models with various masks, in order to improve inpainting performance~\cite{inpaint_attention, irregular, lama}.
For example, LaMa~\cite{lama} succeeded in filling in broad missing areas by extracting periodic structures with Fourier transform, by training with large masks.
On the other hand, several researches focus on additionally tuning the region of a base mask annotated by a user, by using refinements such as dilation~\cite{mask_opt1, mask_opt2}.
Recent models are generally heavy, and the resources required for training have been increasing.
From this aspect, the latter approach, that is mask tuning, will become promising in the era of generative AI, similarly to visual prompting~\cite{visual1, visual2, visual3} for Large Language Models.

In the field of mask tuning, there is little research on what specific shape is desirable for masks.
For example, should masking shapes for a character be preferably in a circle or in a square, or a character contour?
Such perceptible shapes will serve as a user-friendly guideline for manual masking.
Thus, this study focuses on optimizing the shape of masks for image inpainting.
To find the best profile of masks, firstly, mask profiles must be parameterized.
Then, their parameters can be fine-tuned through optimization tools.

Particularly in image inpainting, text removal~\cite{tr_nakamura, tr_zhang, tr_tursun1, tr_zdenek, tr_liu1, tr_tursun2, tr_tang, tr_bian, tr_lyu, tr_liu2, saen, tr_wang, diffstr} is important for industry because it is essential for extraction of background designs, protection of privacy and confidential information~\cite{privacy}.
As an application example, translated posters can be created while maintaining the original design.
Furthermore, characters in text have a complex contour among general target objects in image inpainting.
Document images are suitable for verification on masking process due to the high degrees of freedom.
From these points, we try to optimize mask profiles for a text-removal task.

On text removal, few studies have reported on complex and practical document images. Conventionally, text removal has focused on erasing simple scene text which appears in an outdoor photograph. For example, road signs or signboards are commonly pictured.
However, as shown in Fig.~\ref{fig:benchmark_example}, many document images possess long text, such as a scientific article, advertisement poster, restaurant menu, and catalog flyer.
To the best of our knowledge, there is little benchmark data on text removal for such a document image with long text.
In this paper, to tackle these problems, we construct benchmark data for text removal from images including a large amount of text and develop an optimization technique to design mask profiles.
Based on these benchmark data, the mask design is optimized. 
As explained before, the profile has many degrees of freedom, for example, size, curvature, 
convexity, and so on.
We develop a unified theory to model the profile of typical masks simply and continuously. 
Because the profiles are parameterized, they are ideally optimized using methods such as grid search or exhaustive search.
However, these methods suffer from the curse of dimensionality caused by the high degrees of freedom.
In addition, image generation generally requires a long period of time for evaluation.
Thus, we adopt Bayesian optimization~\cite{BO1, BO2} to search for better parameters.

Our study contributes to the following aspects.
Firstly, this work will contribute to the further practical application of text-removal technology because many documents with long text can be handled.
Secondly, optimized shapes of masks can work as guideline for manual masking.

The remainder of this paper is organized as follows. In Section~\ref{sec:preliminaries}, tasks of text removal are detailed and an overview of the related method is given. 
In Section~\ref{sec:benchmark}, a construction flow is explained for benchmark data on text removal from images including a large amount of text. Then, the detail of the data is summarized. 
In Section~\ref{sec:proposed_method}, we derive parameterized mask profiles for text removal and develop optimization flow for them based on Bayesian optimization.
In Section~\ref{sec:exp}, we perform the following numerical calculations. 
The vulnerability of text-removal  accuracy is estimated against mask profile perturbation.
Then, the profile of masks is optimized through the proposed method.
Finally, Section~\ref{sec:conclusion} concludes this paper.

\section{PRELIMINARIES}
\label{sec:preliminaries}

\subsection{TEXT REMOVAL}
\label{sec:text_removal}
Text removal is a task of erasing characters from given images.
Although this is a fundamental problem that has been known for a long time, the advent of generative models such as image generation has recently changed the situation.
Rather than removing character and simply padding, one can now restore the background design by image generation.
Thus, in this study, text removal is regarded as a task that characters are removed and the background design is consistently restored.

Text-removal tasks can be broadly categorized into two types: mask-prompting and mask-prompting-free, as explained in Section~\ref{sec:introduction}.
In any case, how to design mask profiles is important as shown later.
Our objective is to optimize mask profiles. Therefore, the following section describes related research on profile design.

\subsection{RELATED STUDIES}
\label{sec:related_studies}
\Figure[t!](topskip=0pt, botskip=0pt, midskip=0pt)[width=8cm]{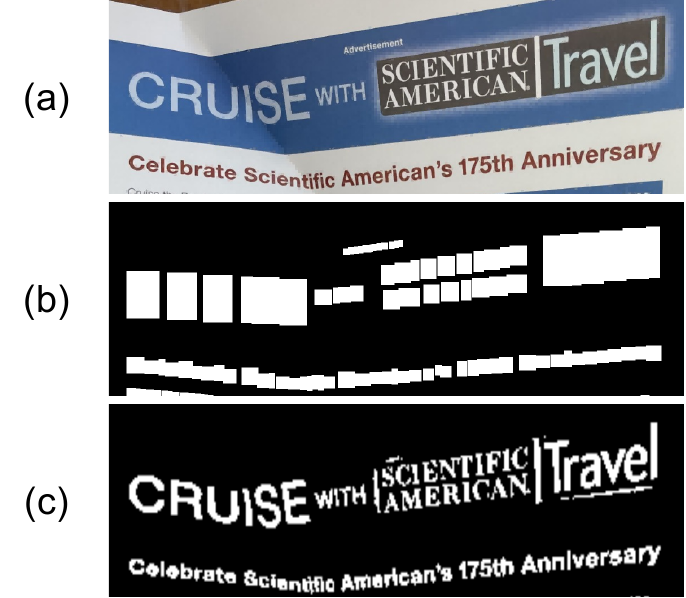}
{Typical masks for text removal. An original image is shown in (a). One of masks is shaped by rough bounds, which is extracted by hand or OCR (b).
The second is defined by pixel-by-pixel (c). Its stroke can be extracted by task-specific model such as SAEN~\cite{saen}.
Type-1 and Type-2 models for mask profiles are based on these base masks.
\label{fig:mask_types}}

There are many researches on enhancing the performance of image inpainting. 
Firstly, several studies have addressed the improvement of robustness against mask variations.
Liu et al. developed the models on irregularly shaped holes with partial convolutions~\cite{irregular}.
LaMa is an image inpainting model that is trained with various mask profiles~\cite{lama}.
Alternatively, several studies focus on refining mask regions by using a dilation process or a segmentation network, in order to assist inpainting networks.
On the other hand, not limited to text removal, there are not many efforts to derive the preffered shape of masks itself, as explained above. 

Typical masks in text-removal tasks are explained.
Firstly, rough bounds for text areas are utilized as a mask~\cite{tr_tursun1, tr_tursun2}.
For example, users manually fill in a square along text, as shown in Fig.~\ref{fig:mask_types} (b).
In place of such manual operations, rough bounds can be automatically obtained using bounding boxes extracetd by Optical Character Recognition (OCR) and image recognition models.
As an advantage of manual operations, users can directly specify the areas to be removed.
The second typical mask is, as shown in Fig.~\ref{fig:mask_types} (c), text strokes defined by pixcel level~\cite{tr_tang, saen, diffstr}.
Bian et al. proposed networks combining both the first and second masks to improve text-removal accuracy~\cite{tr_bian}.
Therefore, there are mainly two approaches: providing broad instructions or specifying details at the pixel level.
This trend is also commonly observed in the field of visual prompting~\cite{visual3}.
As an exception, end-to-end models that bypass intermediate feature masks have also been proposed, which is beyond the scope of this study.

\section{CONSTRUCTION OF BENCHMARK DATA}
\label{sec:benchmark}
Our benchmark is composed of input images and output text-removed images.
As inputs of benchmarks, we downsample $95$ images from open-source OCR datasets~\cite{open1, open2}.
Here, as shown in Fig.~\ref{fig:benchmark_example}, the following three points are visually considered.
\begin{enumerate}
    \item Plentiful text
    \item High OCR difficulty
    \item Diversity in background design
\end{enumerate}
These points reflect the practicality for real-world documents.
Such document images usually contain a text volume significantly greater than that of typical scene-text images.
Physical degradation, such as crumpling and tilting, often occurs due to poor preservation and hand-holding shot, leading to increase of difficulty of OCR~\cite{robust}.
Furthermore, from the practical point of view of image inpainting, a complex design should be considered.
In other words, tasks that can be solved simply by padding based on the paper's base color may be addressed using classical rule-based processing, without the effort on reaching out recent image generation AI. 

Output text-removed images were obtained as follows.
By using image editing tools, texts in input original images were manually erased.
If the process fails to converge in a single trial, this refinement is performed iteratively until the visual discomfort is largely resolved.
Then, super resolution was performed partially to restore the image quality.
Here, IOPaint~\cite{iopaint} and Real-ESRGAN~\cite{realesrgan1, realesrgan2} were used for image editing and super resolution, respectively.
The benchmark data consists of the original images as inputs and the processed image outputs.
Figure~\ref{fig:benchmark_example} shows the examples.
It can be seen that text removal was performed accurately while maintaining consistency in the background design.

\section{PROPSED METHOD}
\label{sec:proposed_method}
In this study, we proposed a method to optimize the profile of masks. First, the specific model is defined by parameterizing mask profiles.  
Then, their parameters are optimized by Bayesian optimization.
In this study, the parameters are called mask parameters.
A surrogate model is trained by data consisting of mask parameters and text-removal scores.
Masks are depicted in an original image according to the predicted mask-parameters. 
After performing image inpainting, we measure the scores of text removal, which are fed back into the existing data.
The surrogate model is updated by re-training these data.
Repeating this process optimizes the profile of masks for a text-removal task.

\subsection{MODELING MASK PROFILE}
\label{sec:modeling}

\Figure[t!](topskip=0pt, botskip=0pt, midskip=0pt)[width=8cm]{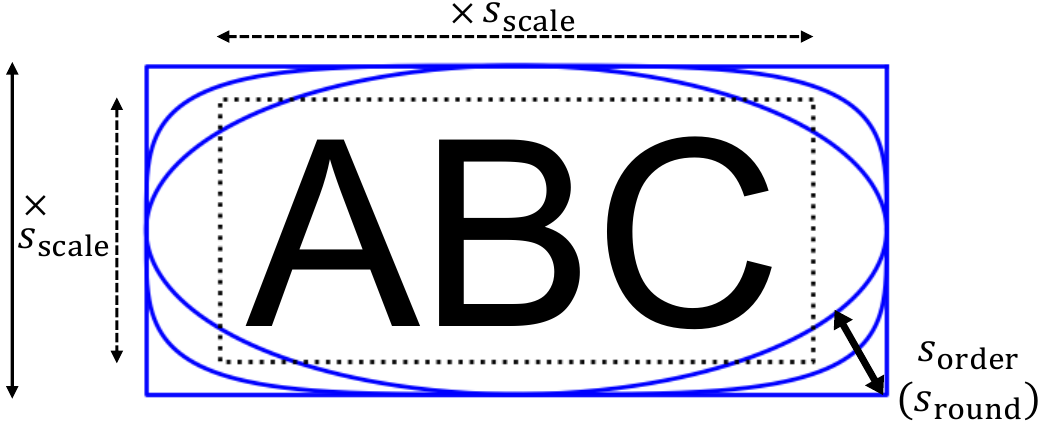}
{Profiles of Type-1 masks are modeled by a superellipse~\eqref{eq:eq3} with a few parameters.
A dotted line shows a base mask, which is originally a pre-detected bounding box for text.
The length of axis is determined by multiplying the box size by $s_{\text{scale}}$. In addition, the roundness is modulated by $s_{\text{order}}$.
\label{fig:line}}
\begin{table*}[t!]
    \caption{Parameters of models for mask profiles are listed. Type-1 is a model based on rough bounds, which are usually extracted by hand or OCR. In our experiment, OCR is used. On the other hand, Type-2 is a model based on pixel-level strokes. One of typical examples is a stroke extracted by SAEN. Base masks are modulated by the following parameters.}
    \label{table:parameters}
        \centering
        \begin{tabular}{ccccccccc}\hline \hline
        \multicolumn{1}{c}{Model} & \multicolumn{3}{c}{Type-1: Eq.~\eqref{eq:eq4}} & & \multicolumn{3}{c}{Type-2: Eq.~\eqref{eq:eq5}} \\
        \cline{2-4}
        \cline{6-8}
        \multicolumn{1}{c}{Parameter} &
             $s_{\text{chunk}}$ & $s_{\text{scale}}$ & $s_{\text{round}}$ & & $t_{\text{thres}}$ & $t_{\text{times}}$ & $t_{\text{kernel}}$ \\ \hline
            Category &Integer &Continuous &Continuous & &Integer &Integer &Integer \\
            Domain  &\{0,1,2\} & [1.0,1.5] &[0.0,1.0] & &\{1, 2,\ldots,100\} &\{-5, -4,\ldots,5\} &\{1,3,5,7\} \\ 
            Descriotion &Chunking option & Enlargement &Roundness & &Contrast &Times of morphology &Kernel size of \\
            & &rate & & &threshold &transformation & morphology tra.\\
            \hline
            &0: character-wise & &0: rectangle & & &$-$: erosion & \\
            Detail &1: word-wise &-- &0-1: superellipse & &-- &$0$: no tra. &-- \\
            &2: paragraph-wise&&1: elipse & & &$+$: dilation & \\\hline
        \end{tabular}
\end{table*}

As explained in Section~\ref{sec:related_studies}, there are mainly two typical masks: rough bounds and pixel-level strokes.
Such mask images are shown in Fig.~\ref{fig:mask_types}.
The typical examples of the former are bounding boxes, which are extracted by hand or OCR.
As degrees of freedom to describe mask profiles, the box size is the most basic variable, which is parameterized by the scale factor $s_{\text{scale}}$.
In other words, letting the size of a base mask $a\times b$, the enlarged mask has the size of $s_{\text{scale}}a \times s_{\text{scale}}b$, as shown in Fig.~\ref{fig:line}.
In this study, the base mask is pre-detected by an OCR model.

In addition, rounding is important because rounded masks are frequently obtained, especially in the case of manually masking.
Thus, the curvature with respect to the bounding box can also be considered a representative variable.
To introduce such a roundness, we utilize the nature of superellipse~\cite{super},
\begin{equation}
\label{eq:eq1}
\left|\frac{2x}{a} \right|^{s_{\text{order}}}+\left|\frac{2y}{b} \right|^{s_{\text{order}}}=1.
\end{equation}
Here, real $s_{\text{order}}$ denotes the order.
While \eqref{eq:eq1} becomes a pure ellipse at $s_{\text{order}}=2$, it converges to a rectangle with $a\times b$ size in $k\to\infty$. this rectangle corresponds to the base mask, as shown in Fig.~\ref{fig:line}.
Figure~\ref{fig:round} shows the results of the obtained mask images.
\Figure[t!](topskip=0pt, botskip=0pt, midskip=0pt)[width=6.5cm]{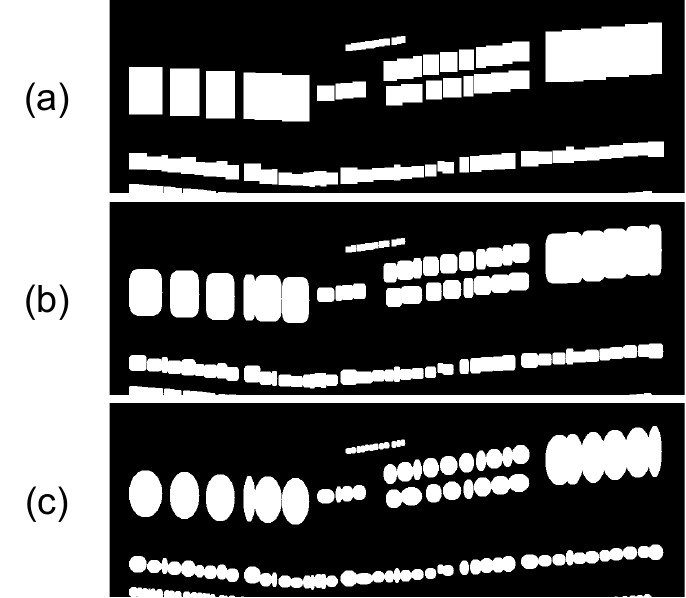}
{Type-1 masks in the shapes of (a) rectangle $s_{\text{round}}=0$, (c) elipse $s_{\text{round}}=1$, and (b) superellipse mask in the middle region.
\label{fig:round}}

There are several options for chunking, i.e. character-wise, word-wise, and sentence-wise, as shown in Fig.~\ref{fig:chunk}.
A categorical variable $s_{\text{chunk}}=\{0,1,2\}$ is introduced for these options, respectively.
\Figure[t!](topskip=0pt, botskip=0pt, midskip=0pt)[width=8.5cm]{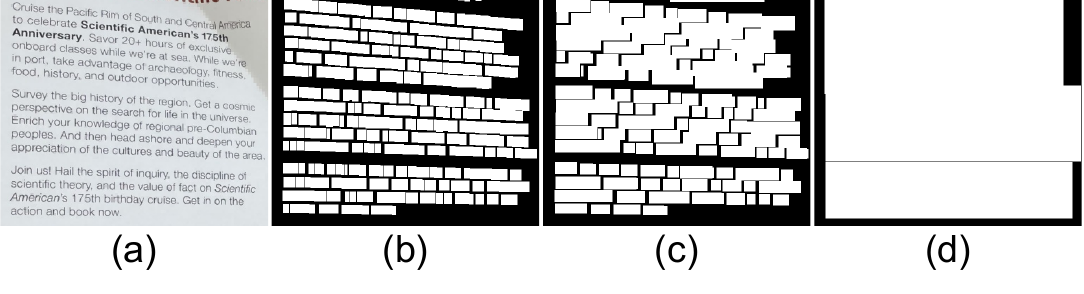}
{Chunk patterns in Type-1 masks. An original image is shown in (a). Type-1 masks with chunking (b) character-wise $s_{\text{chunk}}=0$, (c) word-wise $s_{\text{chunk}}=1$, and (d) paragraph-wise $s_{\text{chunk}}=2$.
\label{fig:chunk}}

Based on the above parameters, the following model (Type-1) is obtained.
\begin{equation}
\label{eq:eq2}
\bigcup_{m\in M_{s_{\text{chunk}}}} \left(\left|\frac{x-x_m}{a_m} \right|^{s_{\text{order}}}+\left|\frac{y-y_m}{b_m} \right|^{s_{\text{order}}}\leq \left(\frac{s_{\text{scale}}}{2}\right)^{s_{\text{order}}}\right).
\end{equation}
Here, $(x_m,y_m)$ and $(a_m,b_m)$ are a center coordinate and size of $m$-th base mask, respectively.
$M_{s_{\text{chunk}}}$ represents the set of base masks, which are predicted for chunks belonging to $s_{\text{chunk}}$ class.
Large parentheses $(\cdot)$ return a bit map. If and only if a proposition in parentheses is true, the bit takes the value of $1$. Otherwise, it takes the value of $0$.
Thus, the above equation~\eqref{eq:eq2} means the bit map for all of the masks. 
By performing an AND operation between this map and original images, masked images are obtained. 

Finally, we define the variables more rigorously.
$s_{\text{chunk}}$ is a ternary variable $\{0,1,2\}$.
In this study, $s_{\text{scale}} \in \mathbb{R}^1$ is ranging from $[1.0,1.5]$.
Considering that line spacing in document images is generally about half the height of a character, the maximum value is set to $1.5$ to prevent overlapping of masks between different lines, as much as possible.
Although $s_{\text{order}} \in \mathbb{R}^1$ originally ranges from $[2,\infty)$, we adpot its reciprocal $s_{\text{round}}=2/s_{\text{order}}$ as a variable because infinite quantities are difficult to handle.
The new variable $s_{\text{round}} \in [0.0,1.0]$ is expected to measure the roundness of a mask.
Thus, for $s_{\text{round}}>0$, Type-1 model is rewritten in
\begin{equation}
\label{eq:eq3}
\bigcup_{m\in M_{s_{\text{chunk}}}} \left(\left|\frac{x-x_m}{a_m} \right|^{\frac{2}{s_{\text{round}}}}+\left|\frac{y-y_m}{b_m} \right|^{\frac{2}{s_{\text{round}}}}\leq \left(\frac{s_{\text{scale}}}{2}\right)^{\frac{2}{s_{\text{round}}}}\right).
\end{equation}
For $s_{\text{round}}=0$, 
\begin{align*}
\label{eq:eq4}
\bigcup_{m\in M_{s_{\text{chunk}}}} 
\left(\left(x\leq x_m+\frac{s_{\text{scale}}a_m}{2}\right) \cap \left(x\geq x_m-\frac{s_{\text{scale}}a_m}{2}\right) \right.   \\ \nonumber
\left. \cap \left(y\leq y_m+\frac{s_{\text{scale}}b_m}{2}\right) \cap \left(y\geq y_m-\frac{s_{\text{scale}}b_m}{2}\right) \right)
\stepcounter{equation}\tag{\theequation} 
\end{align*}
is adopted alternatively.
Type-1 variables $\bm{s}$ are summarized in Table~\ref{table:parameters}.

On the other hand, the typical examples of the pixel-level strokes are determined by the areas of the characters themselves.
The character area can be predicted by task-specific models, such as SAEN~\cite{saen}.
DiffSTR~\cite{diffstr} adopts these base masks and performs dilation on them to obtain broader masks, as shown in Fig.~\ref{fig:dilation} 
Thus, the degrees of freedom to describe mask profiles are the hyperparameters of SAEN and morphology postprocess~\cite{morp}.
The former parameter is a contrast threshold $t_{\text{thres}}$. The SAEN mask is obtained by black-painting the bits in which the color distance between an original image and SAEN-processed image is larger than this threshold.
While DiffSTR considers only erosion in morphology transformation, we additionally utilize dilation.
This is because the SAEN mask may be underestimated when the value of $t_{\text{thres}}$ is large. For example, $t_{\text{thres}}=255$ leads to no masks. 
Thus, in this study, the postprocess is determined by the times of morphology transformation $t_{\text{times}}$ and its kernel size $t_{\text{kernel}}$.
Here, while the positive $t_{\text{times}}$ means a finite number of times of dilation, negative means those of erosion.
\Figure[t!](topskip=0pt, botskip=0pt, midskip=0pt)[width=8cm]{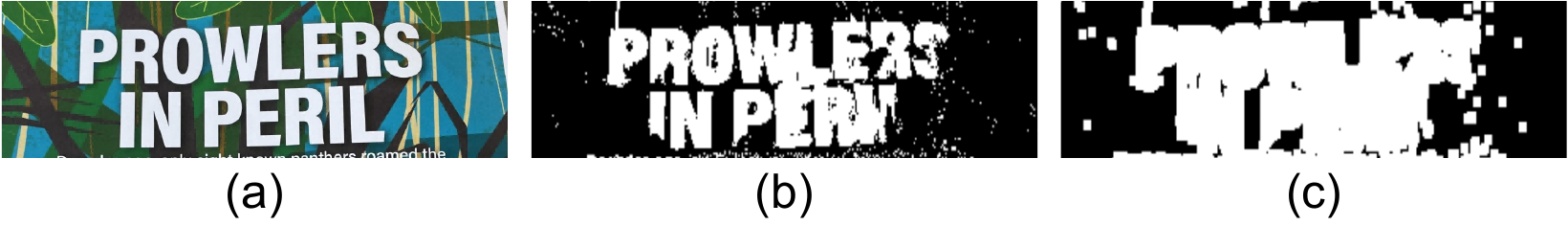}
{Profiles of Type-2 masks are morphologically transformed.
Original image is shown in (a). Base mask obtained by SAEN and dilated mask are listed in (b) and (c).
SAEN has a contrast threshold parameter $t_{\text{thres}}$.
Morphology transformation is parameterized by its times $t_{\text{times}}$ and kernel size $t_{\text{kernel}}$.
\label{fig:dilation}}
\Figure[t!](topskip=0pt, botskip=0pt, midskip=0pt)[width=17.5cm]{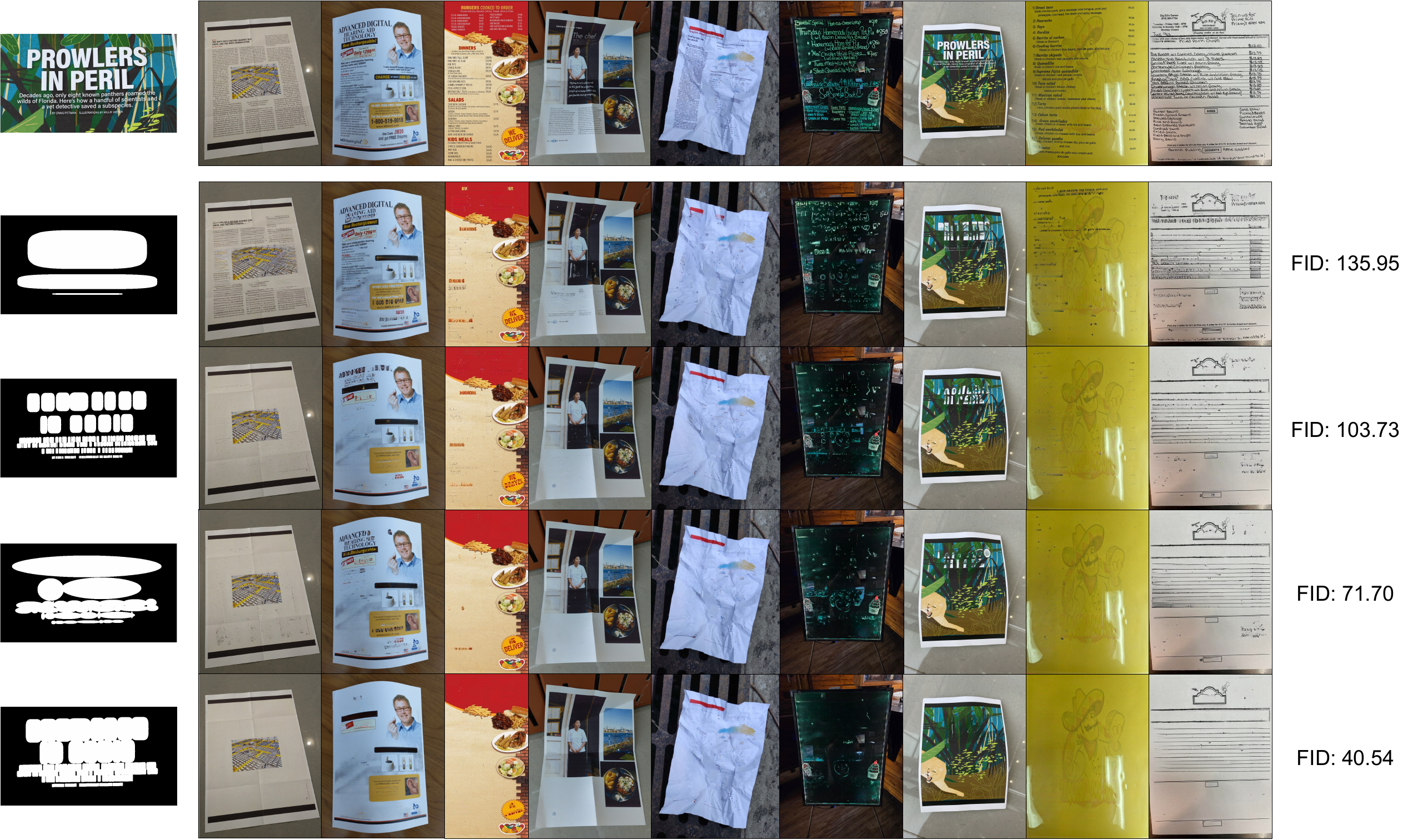}
{Examples of text-removed images and FID scores. Top of figures shows original images.
Below, typical examples are enumerated for each mask profile. They are smapled from initial data.
Left of figures shows mask images and each FID score is in the right.
They are listed in descending order by FID scores.
\label{fig:FID}}

Based on the above parameters, the following model (Type-2) is obtained.
\begin{equation}
\label{eq:eq5}
g(f(I_0,t_{\text{thres}}),t_{\text{times}},t_{\text{kernel}}).
\end{equation}
Here, $I_0$ is an original image and $f(\cdot, t_{\text{thres}})$ returns a bit map processed by SAEN model with the contrast threshold $t_{\text{thres}}$.
$g(\cdot, t_{\text{times}},t_{\text{kernel}})$ returns a final bit map after morphology transformation with the kernel size $t_{\text{kernel}}$ and $t_{\text{times}}$ repetitions.
The image processing of $g$ varies depending on the case:
\begin{equation}
\label{eq:eq6}
g=
\begin{cases}
\text{dilation} & \text{for}\  t_{\text{times}}>0\\
\text{no process} & \text{for}\  t_{\text{times}}=0\\
\text{erosion} & \text{for}\  t_{\text{times}}<0.
\end{cases}
\end{equation}
Type-2 variables $\bm{t}$ are summarized in Table~\ref{table:parameters}.  
$t_{\text{thres}}=1,2,\ldots,100$ and $t_{\text{times}}=-5,-4,\ldots,5$ are integer variables.
In the related study, DiffSTR uses $t_{\text{thres}}=35$ and $t_{\text{times}}=-2$.
Thus, the upper and lower limits are designed by including these preferable values.
$t_{\text{kernel}}$ is a quaternary variable $\{1,3,5,7\}$.
Note that the kernel size is an odd integer by definition, and $t_{\text{kernel}}=1$ means no morphology transformation. 

With respect to the ease of manually reproducing masks, the Type-1 mask has an advantage.
Because the type-2 mask requires tracing characters, users must delicately create masks.
Therefore, using models such as SAEN to automatically determine the type-2 mask is a practical operation for users.

\subsection{OPTIMIZATION FLOW}
\label{sec:flow}
We follow standard Bayesian optimization to optimize mask profiles.
The procedure is a cycle of learning, sampling, and evaluation for data update. 
First, a surrogate model is learned with existing training data.
Their inputs are the parameters of the mask profile models, in other words, $\bm{s}$ in the Type-1 model~\eqref{eq:eq3} and \eqref{eq:eq4} or $\bm{t}$ in the Type-2 model~\eqref{eq:eq5}.
The outputs are the performance score of text removal, as detailed later.
Then, preferable parameters $\bm{s}$ (or $\bm{t}$) for text removal are predicted and sampled.
To update the existing data, the outputs are evaluated with respect to the sampled parameters.
Concretely, the following steps are implemented for inference and evaluation.
The sampled parameters are inputted to the mask profile model, and masked images are generated from original images.
Then, texts are removed from the masked images by using a text-removal model.
The score of text removal is estimated by Frechet Inception Distance (FID)~\cite{fid} between the text-removed image and ground truth.
This score is averaged within our benchmark data.

Usually, the quality of text removal is evaluated with multiple indicators, such as FID, MAE, PSNR, and SSIM~\cite{diffstr}.
However, because this study focuses on single-objective optimization, it is necessary to select one metric.
Therefore, we adopt FID, which is considered the closest to human evaluation~\cite{fid}.

\section{EXPERIMENTS}
\label{sec:exp}
In this chapter, we confirmed that text-removal performance is highly dependent on a mask design, to find that optimizing mask profiles is important for practical application.
Then, we conducted experiments on optimizing mask profiles through Bayesian optimization.

\subsection{VULNERABILITY AGAINST MASK PROFILE}
\label{sec:vulnerability}
If text-removal performance is inherently robust to mask profiles, their parameters are not required to optimize.
To qualitatively confirm this vulnerability, the following experiment was conducted.

By sampling parameters with the grids,
\begin{enumerate}
  \item Type-1
  \begin{enumerate}
    \item $s_{\text{chunk}} = \{0,1,2\}$
    \item $s_{\text{scale}}= \{1, 1.25,1.5\}$
    \item $s_{\text{round}}= \{0, 0.5,1\}$,
  \end{enumerate}
  \item Type-2
  \begin{enumerate}
    \item $t_{\text{thres}}= \{15, 25, 35, 45, 55\}$
    \item $t_{\text{times}}= \{-3,0,3\}$
    \item $t_{\text{kernel}}= \{1,7\}$,
  \end{enumerate}
\end{enumerate}
the test data set was collected. 
Therefore, there are $3\times 3 \times 3 =27$ and $3\times 2 \times 5 =30$ data sets for Type-1 and Type-2, respectively.
Each FID score was estimated as detailed in Section~\ref{sec:flow}.
The data were reused in Section~\ref{sec:optimization} as initial data for Bayesian optimization.
Throughout all experiments in this study, the size of images was resized to $512\times 512$, the processes of masking and morphology transformation were conducted using OpenCV library~\cite{opencv1, opencv2}.
As a base mask for the Type-1 model, we adopted a bounding box extracetd by Google Cloud Vision Api~\cite{cva}.
This api can output each bounding box for chunking options, character-wise, word-wise, and paragraph-wise.
In addition, all reported experiments have been run on Sequoia 15.3 and 3.2 GHz Apple M1 CPU with 16 GB of memory. 

From the Type-1 test data, we selected results with scores of $135.95, 103.73, 71.70, 40.54$, as shown in Fig.~\ref{fig:FID}.
Each result also had entirely different masks.
From the worst score result to the best, the parameters $(s_{\text{chunk}}, s_{\text{scale}}, s_{\text{round}})$ are $(2,1.0,0.5), (0,1.0,0.5), (1,1.25,1.0), (0,1.25,0.5)$. The FID score varies significantly when the mask profile changes.
Among them, the most compact mask had the second worst score.
Thus, masks with the smallest possible area covering character regions are not necessarily optimal.
However, enlarging masks excessively is also inappropriate, as shown in the case of the worst-scored mask.
Although LaMa has been trained to acquire the robustness against mask profiles, our experiment shows that FID scores significantly depended on their design in the case of text-removal tasks.
To alleviate these problems, mask profiles require to be desigined appropriately.

Finally, we remark on the validity of FID scores.
Generally, FID is vulnerable to the number of samples~\cite{cmmd}, and it is recommended to be calculated over a set of ten thousand images or more.
However, our benchmark data falls short of this sample size because acquiring text-removal data takes a significant amount of time due to not only image inpainting but also mask generating. 
Moreover, a multiple number of evaluations are required in black-box optimization, which is the main subject of this study.
Therefore, it is preferable to verify that the small-size FID works reliably as the performance of text removal.
Figure~\ref{fig:FID} shows several text-removed images.
As the FID score decreases, one can see that the results of text removal improve.
The number of remaining characters has decreased, and design consistency has improved.

\subsection{OPTIMIZATION OF MASK PROFILE}
\label{sec:optimization}
\Figure[t!](topskip=0pt, botskip=0pt, midskip=0pt)[width=8cm]{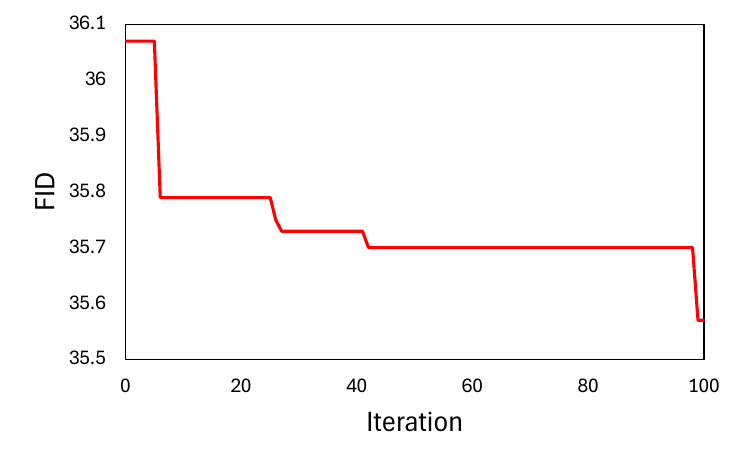}
{FID scores of Type-1 model during optimization process. Best scores at each iteration were plotted.
These lines shall guide the eye. The number of iterations is counted from initial data.
\label{fig:result_type1}}

\Figure[t!](topskip=0pt, botskip=0pt, midskip=0pt)[width=8cm]{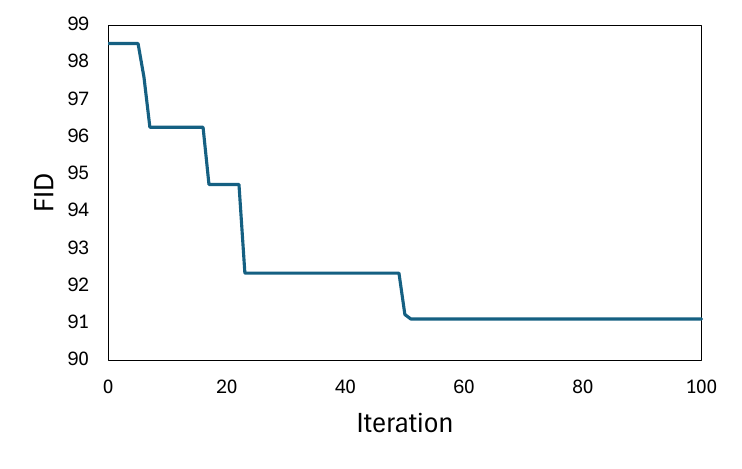}
{FID scores of Type-2 model during optimization process. Best scores at each iteration were plotted.
These lines shall guide the eye. The number of iterations is counted from initial data.
\label{fig:result_type2}}
\Figure[t!](topskip=0pt, botskip=0pt, midskip=0pt)[width=8cm]{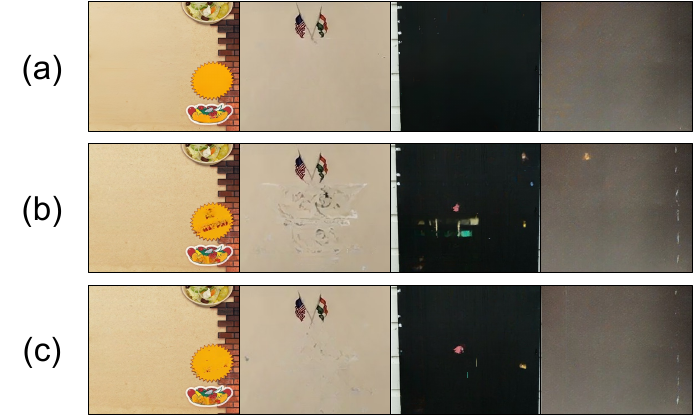}
{Text-removed images during optimization process in Type-1 model, (a) ground truth, processed images with (b) the best FID score in initial data, and (c) the best score in all trials. 
\label{fig:best}}
\Figure[t!](topskip=0pt, botskip=0pt, midskip=0pt)[width=7cm]{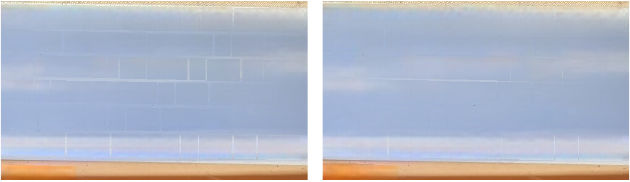}
{Brick-like marks after removing texts. Processed images with the best FID score in initial data and the best score in all trials are shown on the left and right, respectively. While brick-like marks exist before optimization, the optimized masks can largely eliminate such a defect, leading to a homogeneous background.
\label{fig:brick}}

The profile of masks was tuned according to optimization process in Section~\ref{sec:flow}.
As an initial data, we utilized the test data in Section~\ref{sec:vulnerability}.
Then, the optimization process was applied up to maximum iterations, which was set to $100$ in this study.
As a Bayesian optimization solver, Optuna's GPSampler was utilized with default setting~\cite{optuna1, optuna2}.
The best score at each trial is shown in Fig.~\ref{fig:result_type1} and \ref{fig:result_type2}.
Obviously, in both the case of Type-1 and Type-2, the performance of text removal is improved through iterations.
The optimized parameters are $
(s_{\text{chunk}},s_{\text{scale}},s_{\text{round}}
)=(0,1.37,0)$ and $(t_{\text{thres}},t_{\text{times}},t_{\text{kernel}})=(2,7,17)$.
Thus, for the Type-1 model, character-wise masks without roundness were obtained. 
In addition, the area of them was enlarged.
For the Type-2 model, several times of dilation were adopted.
The optimized threshold $t_{\text{thres}}$ was lower than the value reported in DiffSTR~\cite{diffstr}.
In addition, the dilation process was adopted while DiffSTR used eroding.

Text-removed images during optimization process in the Type-1 model are shown in Fig.~\ref{fig:best}. 
Initially, several unremoved residues can be seen in a processed image.
By optimizing the mask parameters, the obtained image is considered to be more faithful to the ground truth.
In addition to mitigation of unremoved residues, brick-like marks were simultaneously reduced, as shown in Fig.~\ref{fig:brick}.
In addition, in the Type-2 model, the performance of text removal is also improved through optimization process.

\subsection{COMPARISON BETWEEN TYPE-1 and TYPE-2}
\label{sec:comparison}
\Figure[t!](topskip=0pt, botskip=0pt, midskip=0pt)[width=7cm]{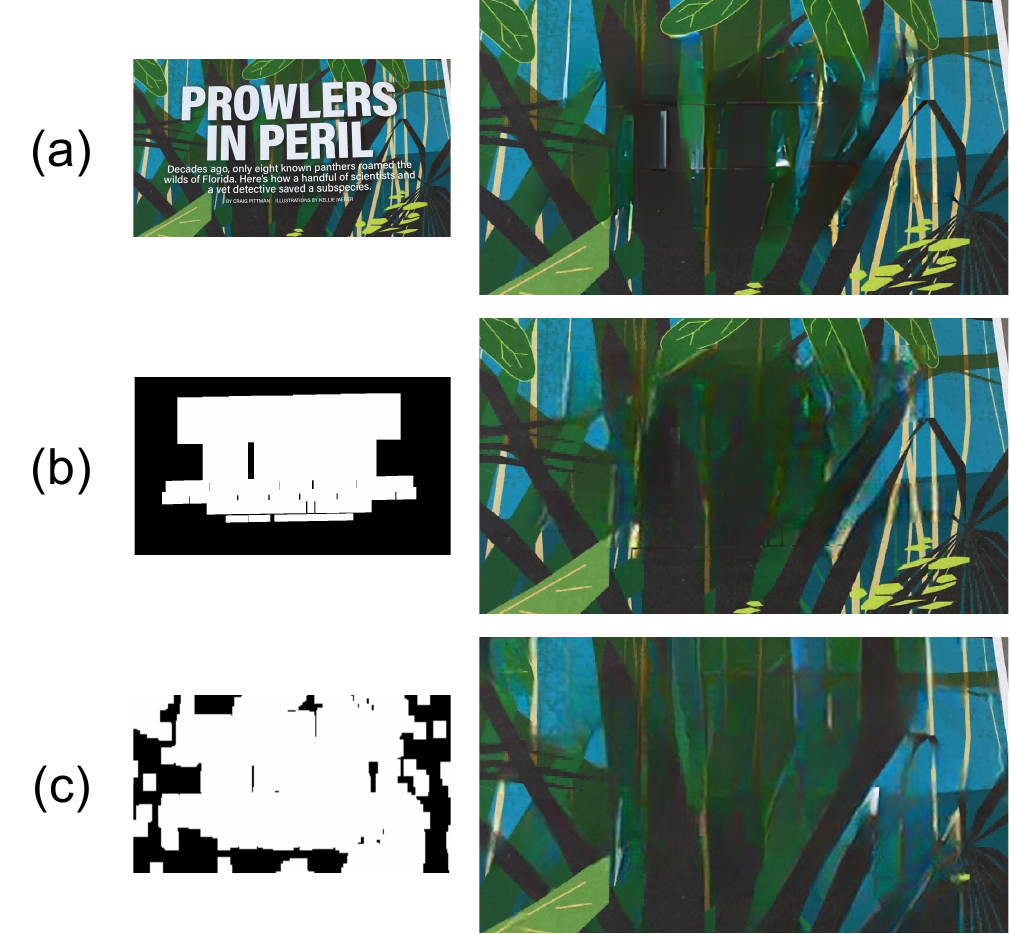}
{Optimized masks and example of text-removed images.
(a) shows an original image (left) and ground truth (right).
Optimized mask profiles for Type-1 and Type-2 model are shown on the left of (b) and (c), respectively. On the right, images inpainted by each mask are shown.
\label{fig:comparison}}
\Figure[t!](topskip=0pt, botskip=0pt, midskip=0pt)[width=7cm]{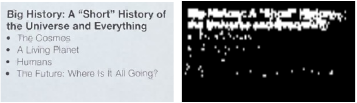}
{Examples of masking failures in SAEN model.
An original image and mask image extracted by SAEN are shown on the left and right, respectively.
For small-font characters, a heatmap becomes patch-like.
\label{fig:failed}}

According to Fig.~\ref{fig:result_type1} and \ref{fig:result_type2}, the Type-1 model shows good scores compared to the Type-2 model.
In fact, as shown in Fig.~\ref{fig:comparison}, the performance of text removal appears to be higher in the Type-2 model.
Because the Type-2 mask has large areas, backgrounds that should not be deleted have been erased.
This may be due to the quality of chracter recognition.
While Cloud Vision Api successfully detected bounding boxes for small characters, SAEN sometimes failed to extract them, as shown in Fig.~\ref{fig:failed}.
Because removing all texts is principally difficult just with such a vanilla mask, it is necessary to expand the mask area as much as possible.
Started from patches in the base mask, the area can be enlarged through dilation operations. The efficiency of enlargement is increased with the kernel size.
Thus, $t_{\text{kernel}}$ is considered to have large values.

In this study, the Type-2 model shows worse scores compared to the Type-1 model.
As a result, it is preferable for users to prompt the character-wise chunked masks with a slightly larger area than an original fontsize, as shown in Fig.~\ref{fig:comparison} (b). 
Users can either automatically create the optimized mask using OCR results as demonstrated in our verification or manually reproduce it as closely as possible.
Note that the profile of the optimized masks can vary depending on which text-removal model is targeted. 
In addition, by fine-tuning SAEN, there is potential for improvement in FID scores of the Type-2 model.

\subsection{PARAMETER DEPENDENCY}
\label{sec:dependency}
\Figure[t!](topskip=0pt, botskip=0pt, midskip=0pt)[width=8cm]{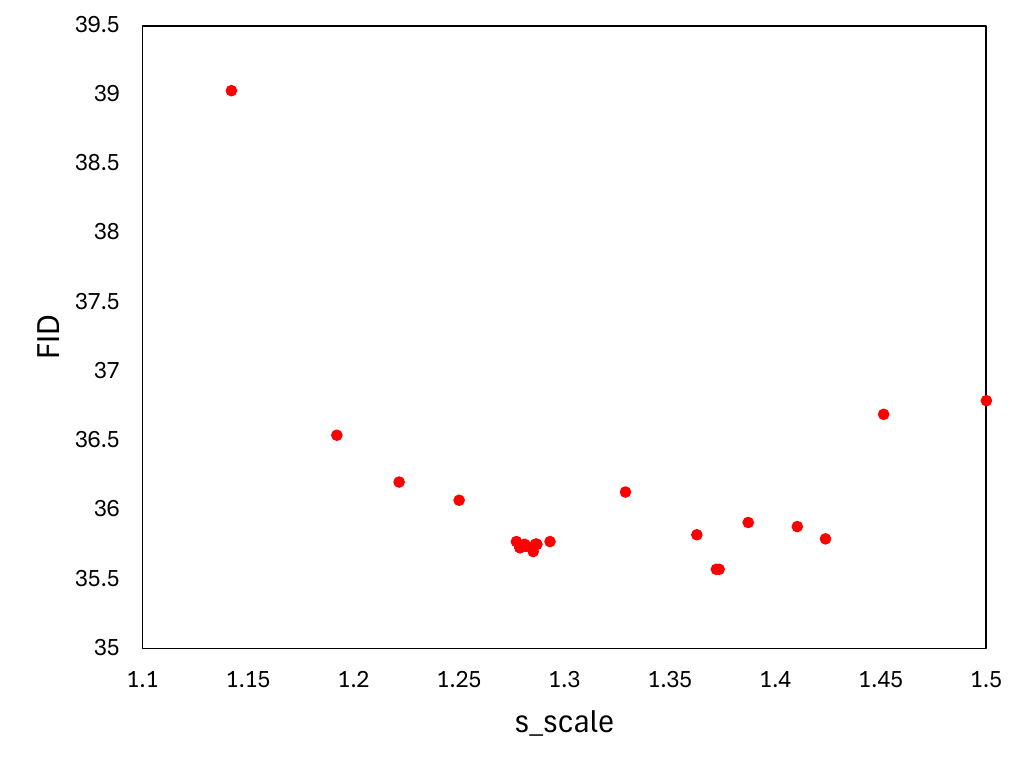}
{Change of Type-1 FID scores with respect to $s_{\text{scale}}$. Here, we plotted partial data where the values of $s_{\text{round}}, s_{\text{chunk}}$ are fixed at $0$, from hystory data through optimization process. 
\label{fig:FID_scale}}
\Figure[t!](topskip=0pt, botskip=0pt, midskip=0pt)[width=7.5cm]{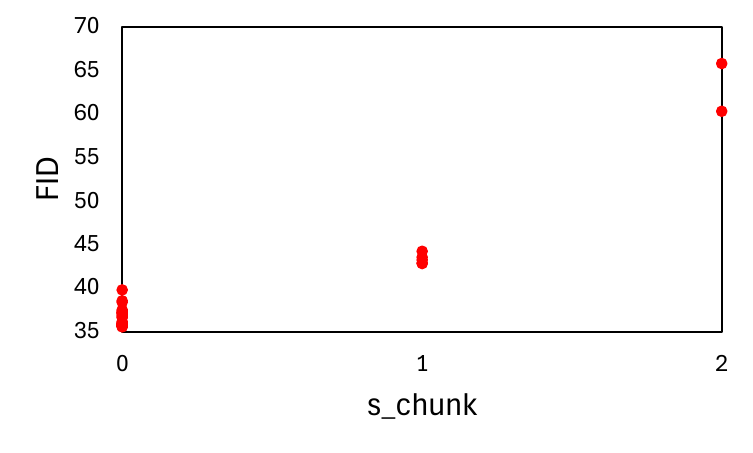}
{Change of Type-1 FID scores with respect to $s_{\text{chunk}}$. 
$s_{\text{chunk}}=0,1,2$ represent character-wise, word-wise, and paragraph-wise respectively.
Here, we plotted partial data where the values of $s_{\text{scale}}, s_{\text{round}}$ are fixed at $[1.25,1.45]$ and $0$, from history data through optimization process. 
\label{fig:FID_chunk}}
\Figure[t!](topskip=0pt, botskip=0pt, midskip=0pt)[width=7.5cm]{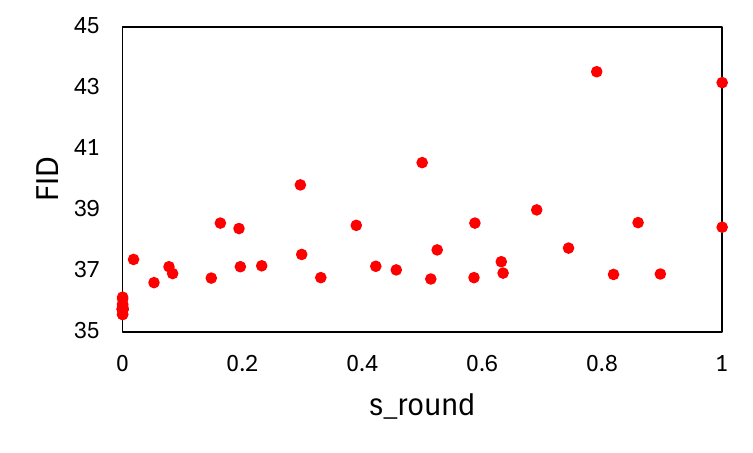}
{Change of Type-1 FID scores with respect to $s_{\text{round}}$.
Here, we plotted partial data where the values of $s_{\text{chunk}}, s_{\text{scale}}$ are fixed at $0$ and $[1.25,1.45]$, from hystory data through optimization process.
\label{fig:FID_round}}
\Figure[t!](topskip=0pt, botskip=0pt, midskip=0pt)[width=7.5cm]{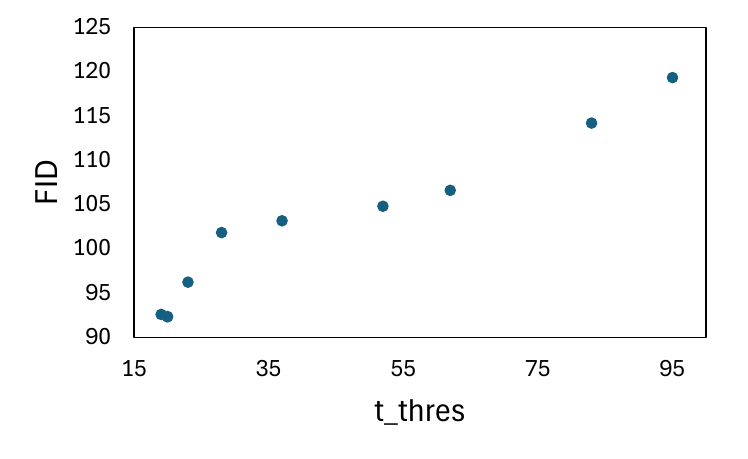}
{Change of Type-2 FID scores with respect to $t_{\text{thres}}$. Here, we plotted partial data where the values of $t_{\text{times}}, t_{\text{kernel}}$ are fixed at $3$ and $5$, from hystory data through optimization process. 
\label{fig:FID_threshold}}
First, we explain Type-1 parameter dependency on FID scores.
Because the optimized scale is higher than $1$, relatively larger masks are preferable.
As shown in Fig.~\ref{fig:brick}, brick-like marks largely disappear. 
Excessively subdividing a target area may enhance the difficulty of cohesive and continuous inpainting, inducing these marks.
Thus, relatively larger masks were expected to be acquired through optimization process.
In fact, around $s_{\text{scale}}=1.37$, the FID scores have a minimum point, as shown in Fig.~\ref{fig:FID_scale}.
This figure plots the Type-1 FID scores with respect to $s_{\text{scale}}$.
Considering that LaMa has been reported to appropriately handle large masks~\cite{lama}, such optimization results are reasonable.

However, masks with $s_{\text{scale}}>1.45$ appear to have reduced accuracy.
This can be considered a natural consequence of the difficulty in repairing excessively large areas.
Figure~\ref{fig:FID_chunk} shows the change of the Type-1 FID scores with respect to $s_{\text{chunk}}$.
Because the FID scores are roughly constant at $s_{\text{scale}}\in [1.25,1.45]$, we plotted partial data within such a domain.
As explained in Fig.~\ref{fig:chunk}, mask areas vary significantly depending on the chunk type.
In other words, the area is amplified as $s_{\text{chunk}}$ increases.
Thus, from Fig.~\ref{fig:FID_chunk}, larger units for chunking show worse scores. 
Especially, in the case of paragraph-wise masks, the FID scores are significantly degraded due to excessive enlargement.

Figure~\ref{fig:FID_round} shows the change of the Type-1 FID scores with respect to $s_{\text{round}}$.
Here, the value of $s_{\text{chunk}}$ is fixed at $0$, and we let $s_{\text{scale}}$ be within $[1.25, 1.45]$.
A tendency toward greater score-dispersion can be observed as roundness increases.
Although several minimum scores are concentrated in $s_{\text{round}}=0$, low scores exist at various roundness values.
This implies that good solutions can be found regardless of roundness.
However, masks with $s_{\text{round}}=0$ show the minimal socres with less dispersion.
Adopting a rectangular shape is a reliable policy for masking.
This result is not trivial because LaMa has been trained not only with boxed masks but also with rounded masks~\cite{lama}.

Next, we explain Type-2 parameter dependency on FID scores.
Figure~\ref{fig:FID_threshold} shows the change of Type-2 FID scores with respect to $t_{\text{thres}}$. Here, we plotted partial data where the values of $t_{\text{times}}, t_{\text{kernel}}$ are fixed at $3$ and $5$.
The FID scores are found to improve as the value of $t_{\text{thres}}$ decreases.
As a result, the larger mask than a base mask has been adopted.
These phenomena are also seen in the case of the Type-1 model.
On the other hand, an excessively large area is also unacceptable.
In fact, the mask with $(t_{\text{thres}},t_{\text{times}},t_{\text{kernel}})=(1,2,7)$ led to the FID score over $300$. 
In principle, appproaching the threshold value $t_{\text{thres}}$ close to zero maximizes the mask area, which means that every pixel becomes masked.
Similarly to the results of the Type-1 model, our process is considered to optimize mask profiles by adjusting their areas.

Unfortunately, compared to the Type-1 model, the model parameters have a lot of levels, leading to the difficulty of independent analysis.
Thus, the dependencies of the other parameters $t_{\text{times}},t_{\text{kernel}}$ are not evaluated.

\section{CONCLUSION}
\label{sec:conclusion}
The advent of generative models has dramatically improved the accuracy of image inpainting. In particular, by removing specific text from document images, reconstructing original images is extremely important for industrial applications. 
Most existing text removers focus on deleting simple scene text which appears in images captured by a camera in an outdoor environment. 
In this study, we address text removing from complex and practical images with dense text. 

We created benchmark data for text removal from images including a large amount of text. From the data, we found that text-removal performance becomes vulnerable against mask profile perturbation.
Thus, for practical text-removal tasks, precise tuning of the mask shape is essential. 
In addition, FID scores were confirmed to effectively reflect text-removal performance.

This study developed a method to model highly flexible mask profiles and learn their parameters using Bayesian optimization. 
Through optimization, FID scores decreased and a drastic improvement in text-removal performance was observed.
The optimized profiles were found to be character-wise masks. It was also found that the minimum cover of a text region is not optimal. 
As a result, brick-like marks after removing text were suppressed and an image consistency is improved.
This result suggests that it is preferable for users to prompt the character-wise chunked masks with a slightly larger area than an original fontsize, as shown in Fig.~\ref{fig:comparison} (b). 
Users can either automatically create the mask based on OCR bounding boxes or manually reproduce it as closely as possible.

Future issues are discussed. 
One drawback of our method is that we did not care for the rotation of bounding boxes, in the Type-1 model.
As shown in Fig.~\ref{fig:benchmark_example}, document images are sometimes rotated.
If Bayesian optimization is applied to masks along rotated boxes, FID scores are expected to improve. 
In addition, optimization with other metrics instead of FID is an interesting study.
Using an improved metric such as CMMD~\cite{cmmd} would also improve text-removal performance.
Besides this performance, multi-objective optimization can take other factors into account, such as processing speed.
Image generation generally suffers from long latency and high cost, which makes accelerating speed and lightweighting models important for practical applications~\cite{syftr}.
Such auto-optimization is realized by parameterizing mask profiles as functions, as detailed in Section~\ref{sec:modeling}.
Finally, we hope that quantitative research on visual prompting will continue to advance.
Current foundation models are extremely heavy, and a trend is emerging to improve their performance by refining prompts without modifying the models themselves.

\appendices

\section*{Acknowledgment}
We wish to express our gratitude to Takashi Egami and Tatsuro Murata. They gave a lot of technical advice on text removal and image inpainting.
We also thank Yoshiyasu Tanaka, Kango Matsushima and Rinka Fukuji for discussion, and Masakazu Yakushiji for research direction.

\begin{IEEEbiography}[{\includegraphics[width=1in,height=1.25in,clip,keepaspectratio]{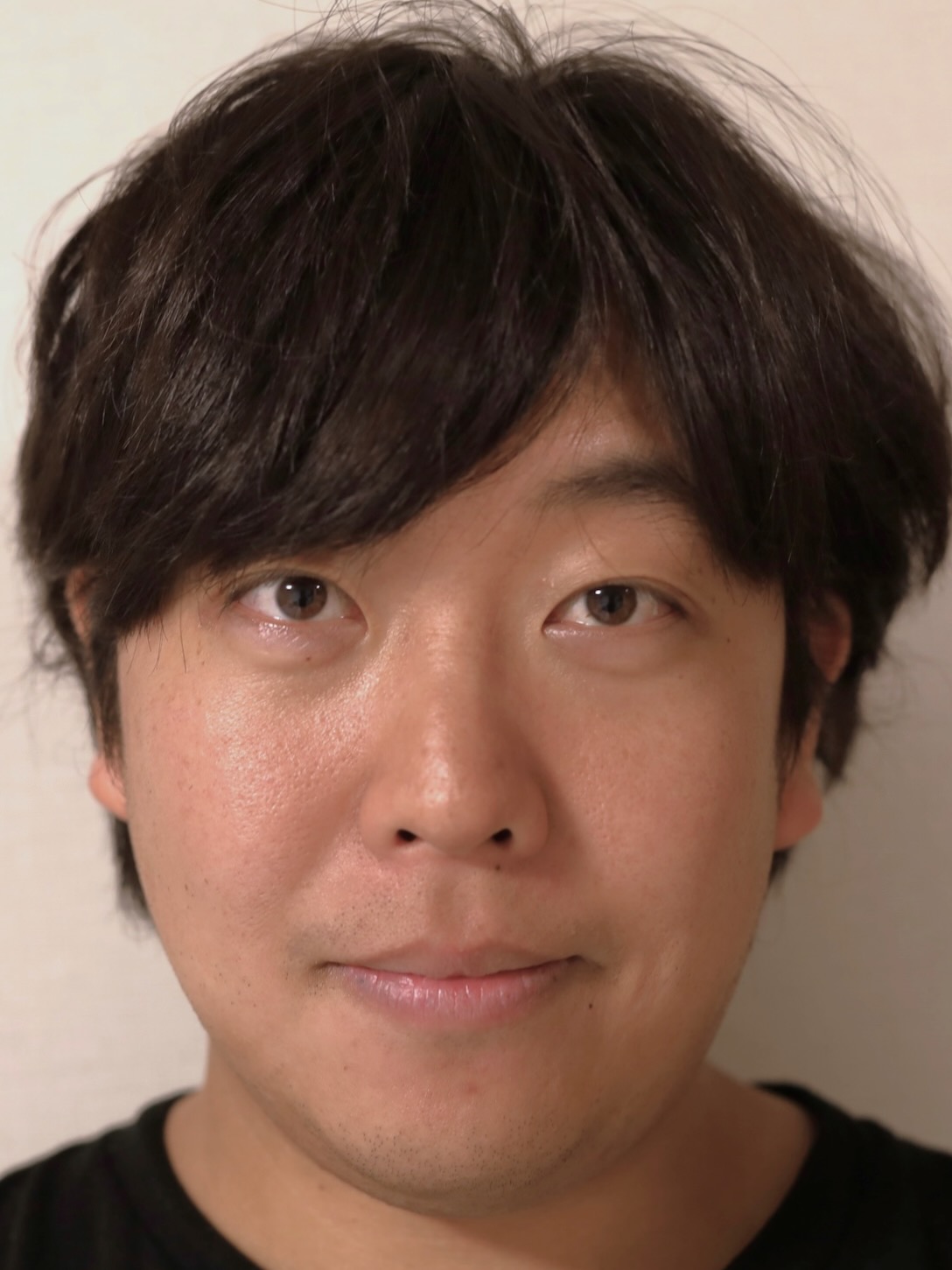}}]{Hyakka Nakada} received his B. Sci. and M.
Sci. degrees from The University of
Tokyo in 2014 and 2016, respectively. 
He is currently pursuing a Ph.~D. degree in applied physics at Keio University.
He is also working for Recruit Co., Ltd., Tokyo, Japan. 
His research interests
include machine learning, optical character recognition, quantum computing, and statistical mechanics.
He is a member of
the Physical Society of Japan (JPS), and The Japan Statistical Society (JSS).
\end{IEEEbiography}

\begin{IEEEbiography}[{\includegraphics[width=1in,height=1.25in,clip,keepaspectratio]{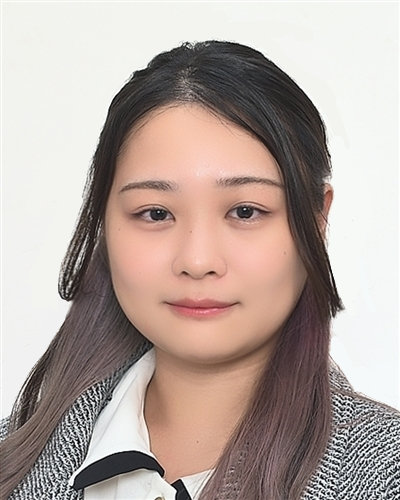}}]{Marika Kubota} received her B. Sci. degree from The University of Ryukyu in 2021.
She is currently with Beans Labo Co., Ltd., Okinawa, Japan.
Her research interests include machine learning.
\end{IEEEbiography}

\EOD

\end{document}